\documentclass[letterpaper, 10 pt, conference]{ieeeconf} 
\usepackage{subcaption}
\usepackage{graphicx}
\usepackage{changepage}
\usepackage{booktabs}
\usepackage{afterpage}
\usepackage{balance}
\usepackage{multirow}
\usepackage[margin=0.769in]{geometry}

\IEEEoverridecommandlockouts 

\usepackage{amsmath} 

\title{\LARGE \bf
Dynamic Subgoal based Path Formation and Task Allocation: A NeuroFleets Approach to Scalable Swarm Robotics
}

\author{Robinroy Peter\textsuperscript{1,†}, Lavanya Ratnabala\textsuperscript{2,†}, Eugene Yugarajah Andrew Charles\textsuperscript{3}, and Dzmitry Tsetserukou\textsuperscript{2}
\thanks{\textsuperscript{†}The authors contributed equally to this work.}
\thanks{\textsuperscript{1}The author is with NeuroFleets (PVT) LTD, Jaffna, Sri Lanka.}
\thanks{\textsuperscript{2}The authors are with the Intelligent Space Robotics Laboratory, Skolkovo Institute of Science and Technology, Bolshoy Boulevard 30, bld. 1, 121205, Moscow, Russia.}
\thanks{\textsuperscript{3}The author is with the Department of Computer Science, University of Jaffna, Jaffna, Sri Lanka.}
\thanks{\tt\ robinroy.peter@neurofleets.com,\{\tt lavanya.ratnabala, d.tsetserukou\}@skoltech.ru, charles.ey@univ.jfn.ac.lk}  
}

\begin{document}

\maketitle
\thispagestyle{empty}
\pagestyle{empty}

\begin{abstract}
This paper addresses the challenges of exploration and navigation in unknown environments from the perspective of evolutionary swarm robotics. A key focus is on path formation, which is essential for enabling cooperative swarm robots to navigate effectively. We designed the task allocation and path formation process based on a finite state machine, ensuring systematic decision-making and efficient state transitions. The approach is decentralized, allowing each robot to make decisions independently based on local information, which enhances scalability and robustness. We present a novel subgoal-based path formation method that establishes paths between locations by leveraging visually connected subgoals. Simulation experiments conducted in the Argos simulator show that this method successfully forms paths in the majority of trials. However, inter-collision (traffic) among numerous robots during path formation can negatively impact performance. To address this issue, we propose a task allocation strategy that uses local communication protocols and light signal-based communication to manage robot deployment. This strategy assesses the distance between points and determines the optimal number of robots needed for the path formation task, thereby reducing unnecessary exploration and traffic congestion. The performance of both the subgoal-based path formation method and the task allocation strategy is evaluated by comparing the path length, time, and resource usage against the A* algorithm. Simulation results demonstrate the effectiveness of our approach, highlighting its scalability, robustness, and fault tolerance. 
\end{abstract}

\section{Introduction}
Robotics has rapidly grown, with multi-robot systems (MRS) becoming essential for handling complex tasks. These systems involve multiple robots working together to achieve common goals, but coordinating them presents challenges, especially in terms of autonomy and human factors. Swarm robotics, inspired by social animals like ants and bees, uses simple robots with limited sensing and communication capabilities that interact locally and self-organize to achieve global behavior, ensuring scalability, adaptability, and reduced costs.
\begin{figure}[ht]
 \centering
 \includegraphics[width=0.4\textwidth]{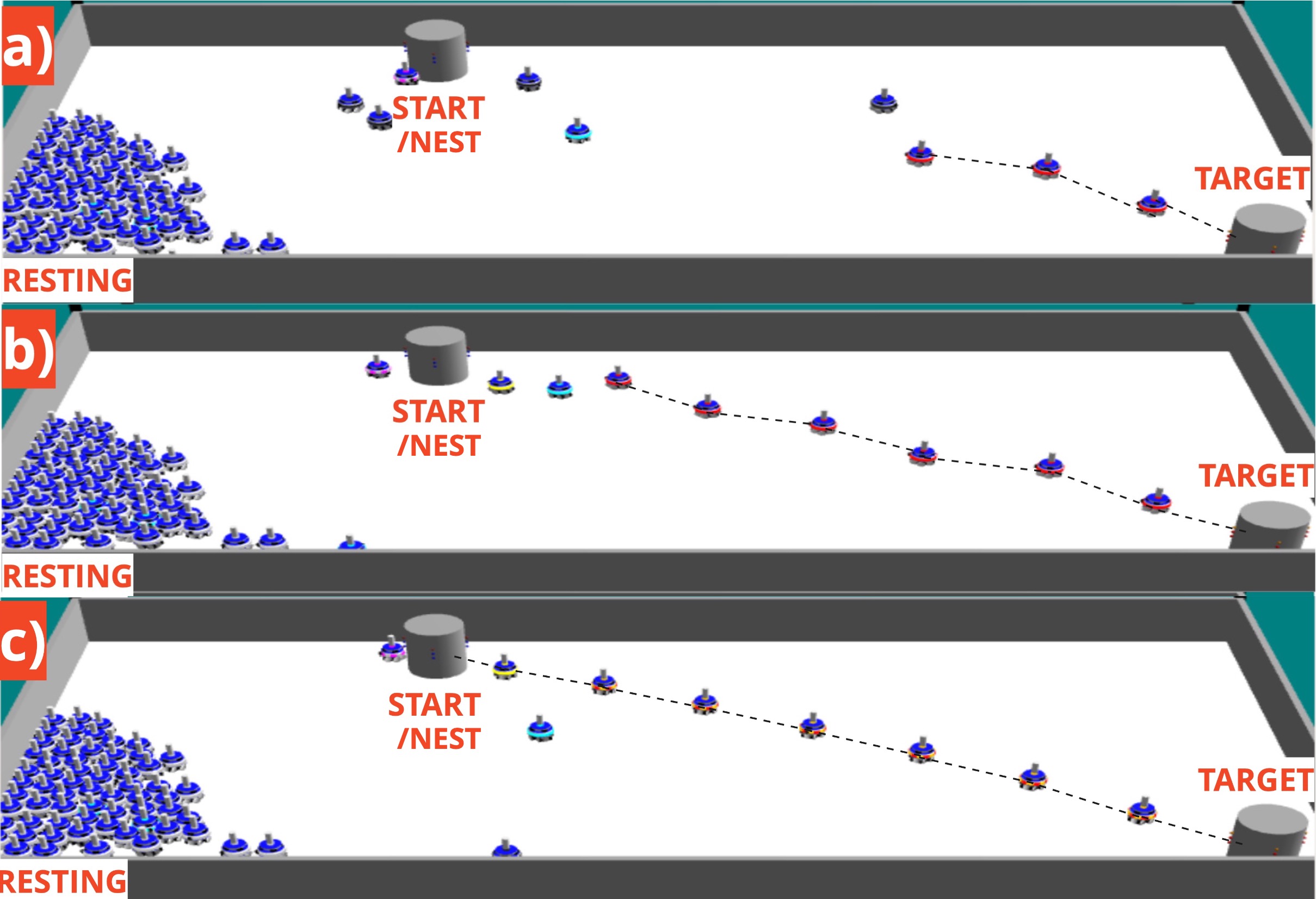}
 \caption{Illustration of dynamic swarm pathway creation in the Argos simulator: (a) Formation of three subgoals. (b) Completion of subgoal paths. (c) Final optimized path after two heuristic optimization steps.}
 \label{fig:test}
  \vspace{-0.5cm}
\end{figure}
This paper focuses on subgoal-based path formation for swarms of robots with limited sensing and visual communication capabilities. Robots are able to create a path between a start and an end point through the use of visually accessible subgoals and light signal-based interactions. Yet, using a large number of robots might result in traffic congestion, robot-to-robot collisions, and decreased efficiency.

To address this, we propose a task allocation strategy that optimizes robot deployment. This strategy allows only the necessary number of robots to perform the path formation task, while others are assigned to rest, reducing collisions and congestion. Fig.~\ref{fig:test} shows the Argos simulation environment, which effectively generates rr-bot behavior and tests the subgoal-based path formation method. This method can be applied to various tasks, enhancing overall system efficiency and performance.

\section{Related Works}
Collective navigation involves robots reaching a destination by traversing unknown environments with assistance from other robots using communication techniques and finite-state machines. One study proposed a strategy for transporting large objects using numerous mobile robots, which push the object at points where the goal's direct line of sight is obstructed \cite{b1}. In order to navigate complex surroundings, it was also advised to develop a subgoal-based path construction technique by placing intermediate goals in between the start and end points. Robots follow the path, pushing the object towards each subgoal until reaching the final goal. Another strategy involved negotiating goal direction for transportation of large objects \cite{b17}. Inspired by ant foraging, researchers proposed methods for creating efficient paths using artificial pheromones \cite{b22}, employing techniques like releasing alcohol, heat, odor, visual marks, or RFID tags. However, these systems may not be reliable in realistic scenarios, prompting the proposal of local IR range bearing as an alternative. Efficient path formation can also be achieved using field-based methods, such as vector-field and chain controllers \cite{b3}. Robots form paths based on LED light directions, deciding probabilistically whether to join the path. Evolutionary-based approaches have also shown promise in generating the paths between targets by solving cooperative tasks in swarm robotics \cite{b2}. Authors of \cite{b8} proposed Deep Reinforcement Learning (DRL) based path planning for 3-dimensional drone landing tasks.  Authors of \cite{b23}, \cite{b24} used Deep learning for navigation.

Various strategies exist for task allocation in swarm robots. In multi-foraging scenarios, the Distributed Bees' Algorithm (DBA) \cite{b10} uses broadcast communication to inform robots of target locations. Other methods rely solely on local interactions, eliminating the need for global knowledge and centralized components \cite{b11}, \cite{b4}, \cite{b5}. Another approach \cite{b12} employs task partitioning using sigmoid functions, while the response threshold sigmoid model helps avoid traffic congestion \cite{b19}. Novel approaches assign robots to tasks using simple reactive mechanisms or advanced gossip-based communication \cite{b13}. Hierarchical task assignment and path finding for limited communication swarms have also been proposed \cite{b7}, along with the methods based on optimal mass transport theory \cite{b6}. Self-organizing approaches allocate robots for foraging tasks with sequential sub-tasks based on response thresholds \cite{b18}, while other algorithms allocate homogeneous robots to heterogeneous tasks \cite{b20}, \cite{b21}. 

The Argos simulator \cite{b16} is a notable tool for swarm robotics research, capable of simulating up to 10,000 robots simultaneously. It allows different physics engines to be applied to various regions of the arena, making it highly suitable for swarm robotics simulations.

\section{Methodology}
In this study, we propose a two-step approach to address path formation and task allocation challenges in swarm robotics. First, we implement a task allocation strategy to prevent traffic congestion and collisions, determining the optimal number of robots needed for path formation while assigning the rest to rest. Then, we use a subgoal-based method to establish paths between the start and goal points.
\subsection{Environment Setup}
\begin{figure}[h]
\centering
 \begin{subfigure}[b]{0.15\textwidth}
    \includegraphics[width=\textwidth]{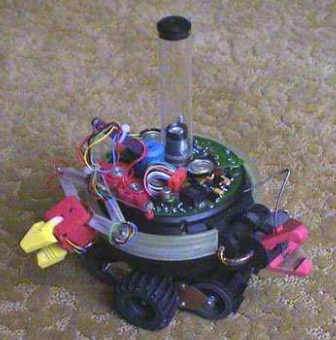}
    \caption{}
    \label{fig:sbot1}
 \end{subfigure}
 \hspace{15pt} 
 \begin{subfigure}[b]{0.15\textwidth}
    \includegraphics[width=\textwidth]{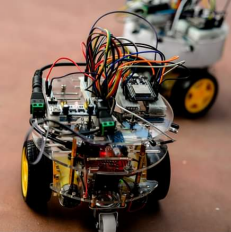}
    \caption{}
    \label{fig:sbot2}
 \end{subfigure}
 \caption{Physical robot. (a) S-bot. (b) Custom design physical rr-bot.}
  \label{fig:sbot}
\end{figure}

The environmental setup for this study revolves around the s-bot (see Fig.~\ref{fig:sbot1}), developed as part of the SWARM-BOTS project. While individual s-bots have limited capabilities, a swarm of s-bots can overcome these limitations and operate efficiently. Since our own physical rr-bots (see Fig.~\ref{fig:sbot2}) are still under construction, all experiments were conducted using the Argos simulation software. The simulation includes Artificial Potential Fields (APF) towards the starting point, with robots having repulsion and attraction fields to navigate efficiently. In large-scale deployments, collisions among robots are minimized using a diffusion vector. This vector directs the robots' movements after a collision, guiding them away from congested areas to prevent further collisions. This approach ensures smooth operation and reduces congestion within the swarm. Fig.~\ref{fig:sbot} shows the physical robot setup for testing. 

We also set up the simulation environment with specific parameters that govern the behavior of the robots (see Fig.~\ref{fig:test}).
The diffusion parameters control the movement characteristics of the robots, particularly in how they navigate through the environment. State parameters define the different states a robot can be in during the simulation, such as resting or exploring. The wheel turning parameters control the maneuverability of the robots, particularly how sharply they can turn. These parameters used in our simulations are listed in Table~\ref{tab:combined_parameters}.

\begin{table}[h!]
    \centering
    \caption{Simulation Parameters.}
    \begin{tabular}{|l|c|}
        \hline
        \multicolumn{2}{|c|}{\textbf{Diffusion Parameters (deg.)}} \\ \hline
        Go straight angle range & -5 - +5 \\ \hline
        {delta} & 0.1 \\ \hline
        \multicolumn{2}{|c|}{\textbf{State Parameters (s)}} \\ \hline
        Minimum Resting time & 0.1  \\ \hline
        Initial Exploring time & 1 \\ \hline
        Minimum search for place in nest & 5 \\ \hline
        \multicolumn{2}{|c|}{\textbf{Wheel Turning Parameters}} \\ \hline
        Hard turn angle threshold (deg.) & 90 \\ \hline
        Soft turn angle threshold (deg.) & 70 \\ \hline
        No turn angle threshold (deg.) & 10 \\ \hline
        Maximum speed (m/s) & 10 \\ \hline
    \end{tabular}
    \label{tab:combined_parameters}
\end{table}

\begin{figure}[h]
    \centering
    \begin{subfigure}[b]{0.17\textwidth }
        \includegraphics[width=\textwidth]{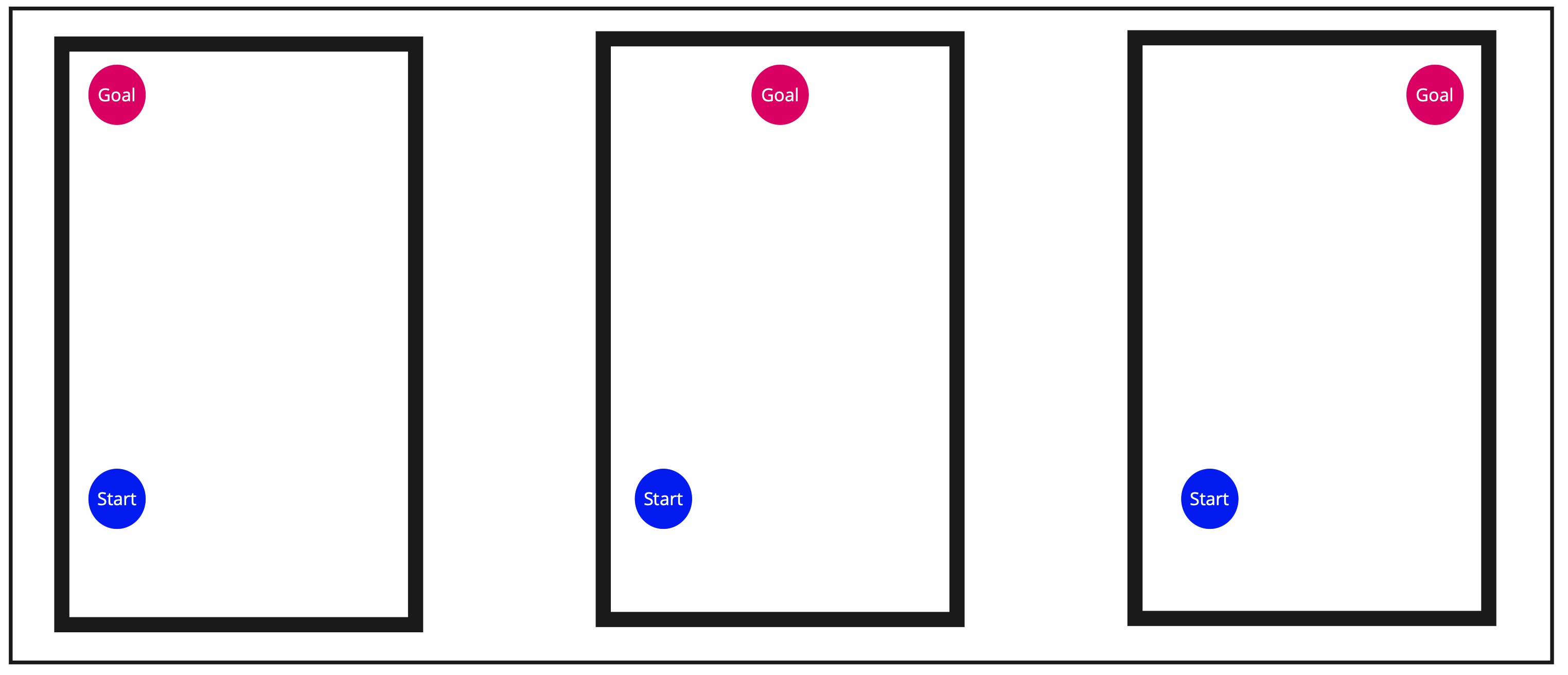}
        \caption{}
        \label{fig:env1}
    \end{subfigure}
    \hfill
    \begin{subfigure}[b]{0.17\textwidth}
        \includegraphics[width=\textwidth]{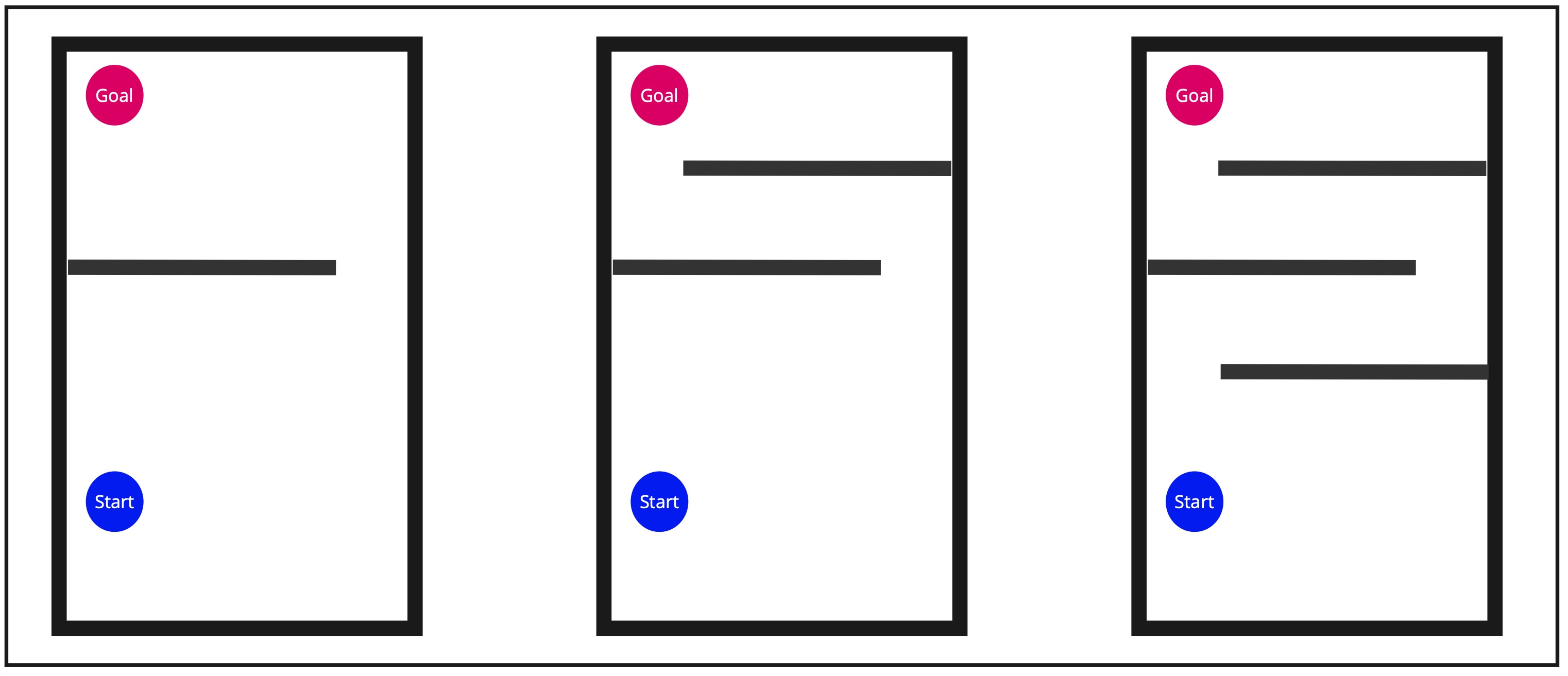}
        \caption{}
        \label{fig:env2}
    \end{subfigure}
    \hfill
    \begin{subfigure}[b]{0.11\textwidth}
        \includegraphics[width=\textwidth]{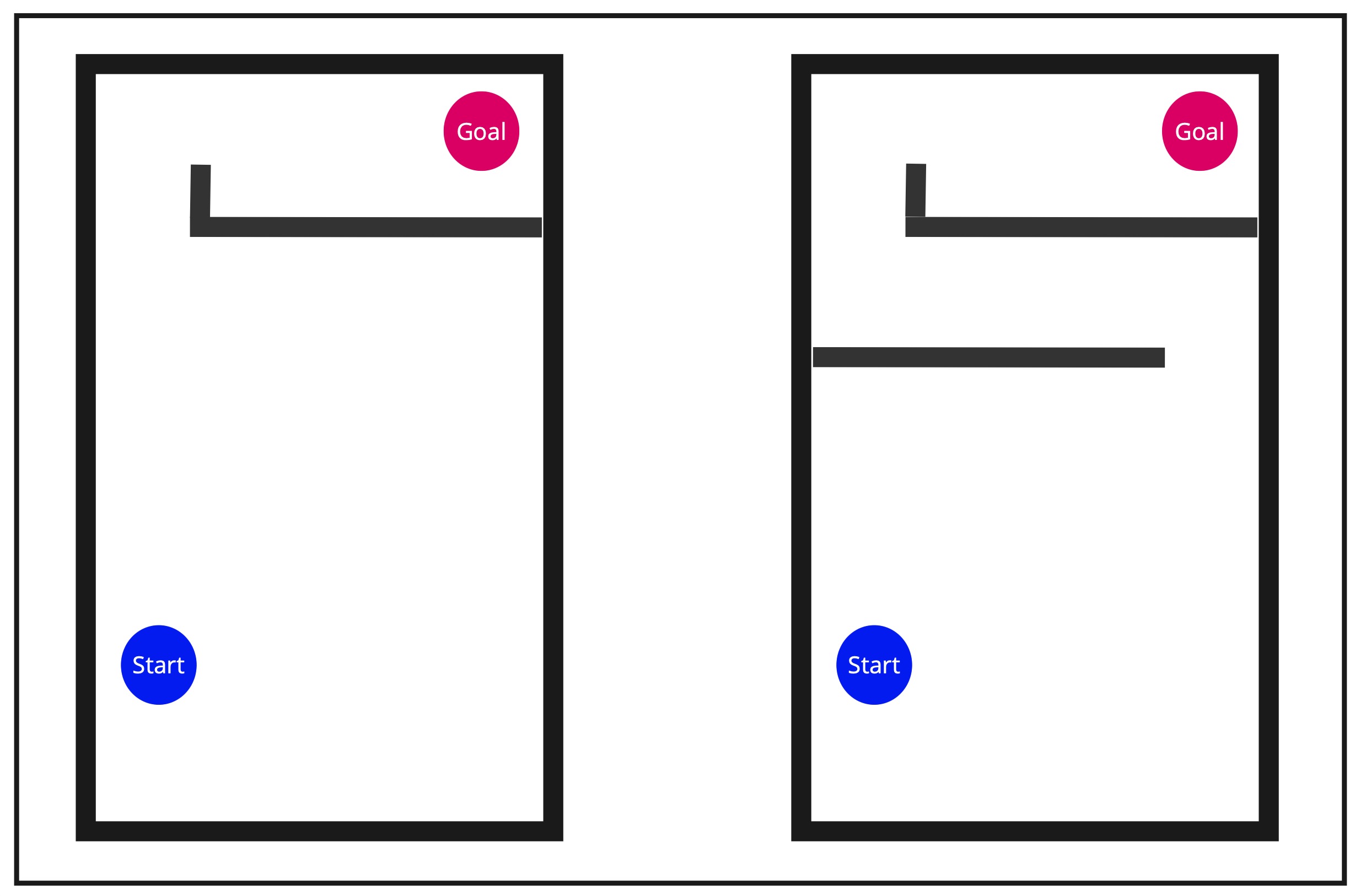}
        \caption{}
        \label{fig:env3}
    \end{subfigure}
    \caption{Robot start (blue) and goal (pink) points in various Argos simulation environments (8m x 4m): (a) Open. (b) Obstacle. (c) Complex obstacle.}

    \label{fig:env_all}
    \vspace{-0.3cm}
\end{figure}
\subsection{Task Allocation Strategy}
We implement a task allocation strategy before initiating the path formation task to avoid inter-robot collisions, traffic congestion, and unnecessary exploration time and resource usage, with the primary aim of minimizing the operational costs. Initially, all robots are deployed into the environment to begin exploration. Each robot explores the environment until it reaches its maximum exploration time or depletes its energy, at which point it returns to the starting point to recharge. During the exploration phase, if a robot detects the goal within its visual range, it becomes the “goal founder” and changes its LED color to signal the discovery. This color change alerts other robots that the goal has been found, prompting them to transition to the decision-making state. In this setup, robots in the decision-making state and those encountering them follow a potential field, i.e. a navigational guide similar to a magnetic field, back to the starting point to initiate the task allocation.
The task allocation process involves two main tasks: path formation and resting. The goal founder robot calculates the path length using its exploration time and speed. Based on this path length and the robots' visual communication range, the goal founder determines the number of robots required for the path formation task:
\begin{equation}
n=s*t/v+\delta ,
\end{equation}
where $n$ is the number of needed robots for path formation task, $s$ is the speed of the robot, $t$ is the exploration time, $v$ is the visual range and $\delta$ is complexity factor.

\begin{figure}[h]
 \centering
 \includegraphics[width=0.4\textwidth]{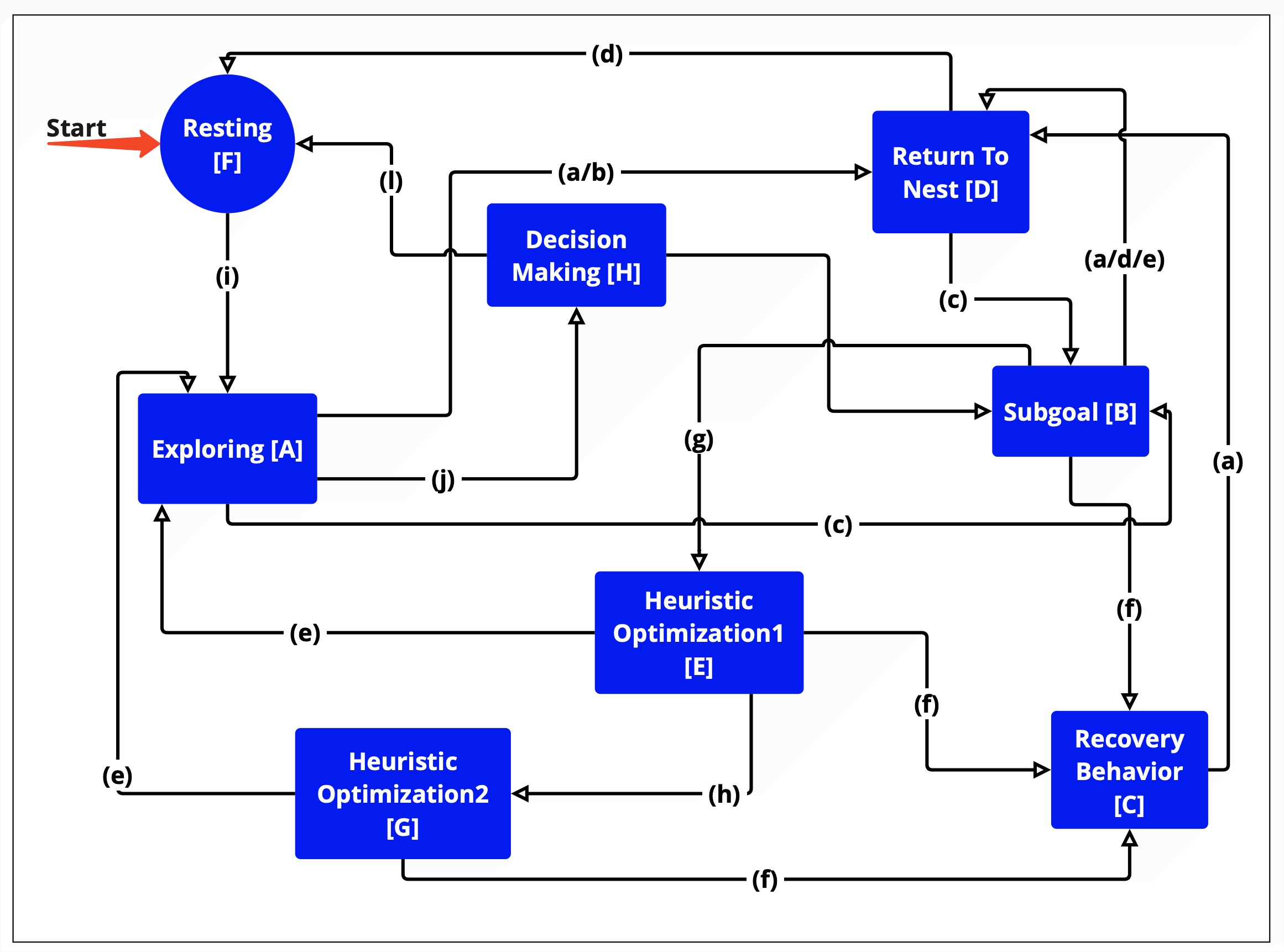}
 \caption{Finite state machine diagram for subgoal-driven path formation with task-allocation.}
 \label{fig:fsm}
 \vspace{-0.3cm}
\end{figure}

\begin{figure}[ht]
 \centering
 \includegraphics[width=0.4\textwidth]{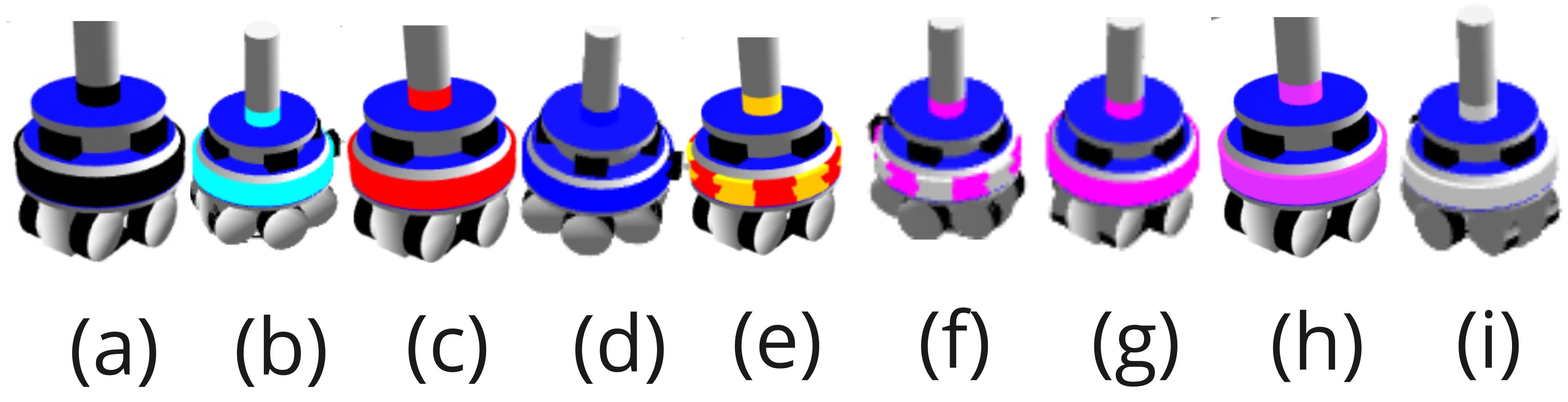}
 \caption{Color dynamics of a robot across various operational states: (a) Exploring (black). (b) Return to nest (cyan). (c) Subgoal (red). (d) 1\textsuperscript{st} Optimization (blue), (e) 2\textsuperscript{nd} Optimization (red-yellow). (f) Goal founder (dashed magenta). (g) Recovery (magenta). (h) Decision-making (intensive magenta). (i) Resting (white).}
 \label{fig:robotstate}
  \vspace{-0.5cm}
\end{figure}

Throughout this process, light-signal-based interactions were used to find out the needed robot count. The finite state machine model depicted in Fig.~\ref{fig:fsm}, along with the detailed information provided in Table \ref{tab:StateTable} and Table \ref{tab:Transitions}, serve as the foundation for our subgoal-based path formation method and task allocation process. Fig.~\ref{fig:robotstate} shows the different state colors of the robots. 

\begin{table}[h!]
\caption{Description of Behavioral States.}
    \renewcommand{\arraystretch}{1.2}
	\begin{tabular}{|r|p{1.95in}|} 
	\hline
	States&Description\\
	\hline
		Resting&Resting in place for a set number of steps.\\
		Exploring&Robots search for the goal from the starting point, moving against potential fields. They have a maximum step size and if unable to find the goal, they switch to returning to the nest. This return is guided by an Artificial Potential Field (APF) to ultimately reach the goal.\\
		Subgoal&When a robot identifies a goal or subgoal, it positions itself as a subgoal between the starting point and the detected target. \\
		Return To Nest& Robots start from the nest and search for the goal against the potential field, limited by their maximum step size. If they fail to find the goal, they return to the nest to increase their search radius for the next attempt. \\
        Decision-Making&When a robot discovers a goal or encounters a goal-finding robot, it transitions to a state where it decides, based on local communication protocols, whether to form a path or rest.\\
		Recovery Behavior&When a robot loses sight of a detected subgoal or goal, it shifts to a subgoal forming state. Here, it initiates recovery behavior to keep other robots out of a blind spot within a specified radius. \\
		Heuristic  Optimization1&Once path formation is complete at the starting point, the nearest subgoal robot begins aligning itself in a straight line from the start.\\
		Heuristic  Optimization2&Once the first optimization at the goal is complete, the second optimization begins with the robot nearest to the goal and proceeds from the goal towards the start, mirroring the first optimization.\\
    \hline
	\end{tabular}
	\label{tab:StateTable}
\end{table}
\begin{table}[h!]
\caption{Overview of State Transition Mechanisms.}
    \centering
    \begin{tabular}{|r|p{2.2in}|}
	\hline
	Transitions&Description\\ 
	\hline
		a&Path formation successfully completed.\\
		b& Unsuccessful exploration. \\
		c& Found the goal/subgoal.\\
		d&Reach nest.\\
		e& Found another swarm robot in B, E, G state to archive the end position of that state.\\
		f&Lost visibility of the goal/subgoal within that range.\\
		g&Successfully formed as a subgoal and detected color patterns to transition to state E.\\
		h&Successfully completed Heuristic Optimization1 and detected color patterns to transition to state G.\\
		i&The resting period has ended.\\
  
		j&Once an agent finds a goal, it enters this state and calls other agents to help allocate resources for path formation tasks.\\
		k& After finding the target, the robot returns to the nest to decide how many robots are needed for path formation during the current exploration.\\
        l& Once the necessary robots are assigned to path formation tasks, the remaining robots will engage in resting tasks.\\
	\hline
	\end{tabular}
	\label{tab:Transitions}
\end{table}
\subsection{Local Communication Protocol}
In the proposed communication protocol, the goal-finding robot utilizes broadcast signals to disseminate task requests across all robots within range, while individual robots employ uni-cast signals to transmit specific data such as robot IDs and acknowledgments back to the goal-finding robot. The data array is strategically structured where Data[0] and Data[1] divide the exploring time into two parts due to integer constraints, Data[5] and Data[6] similarly encode the robot ID, Data[7] signifies the termination of task allocation, Data[8] is used for sending acknowledgment responses, and Data[9] facilitates the broadcast request for path formation tasks. Initially, each robot configures its data array with its ID and exploring time. The goal-finding robot then broadcasts a task request, which is acknowledged by interested robots through an unicast response that includes their ID and acknowledgment. This process continues until the optimal number of robots is engaged, at which point a termination message is broadcasted, signaling the remaining robots to either rest or undertake alternative tasks. This system ensures efficient coordination and clear role distribution among the robots, enhancing the collective task execution within the robotic swarms. Based on the communication protocol, tasks $T$ are categorized as:
\begin{equation}
T_i = 
\begin{cases} 
\text{Path Formation} & \text{if } i \leq n \text{ (first } n \text{ robots to respond)}, \\
\text{Resting} & \text{if } i > n,
\end{cases}
\end{equation}
where \( i \) denotes the order in which robots receive and respond to the broadcast signal. Robots that are among the first \( n \) to respond are allocated to the path formation task, while the others are assigned to the resting task.

\subsection{Subgoal-Based Path Formation}
In the subgoal-based path formation phase within the Argos environment, robots start by exploring the area to locate a goal. They engage in random exploration, gradually increasing their distance from the nest. If they fail to find the goal within the minimum set exploration time, they return to the starting point for another attempt. Upon detecting a goal within 30 cm, a robot enters the subgoal state, emits a white color signal, and uses the APF method to head back towards the start, positioning itself as a static subgoal at 70 cm.

\begin{figure}
    \centering
    \includegraphics[width=1.0\columnwidth]{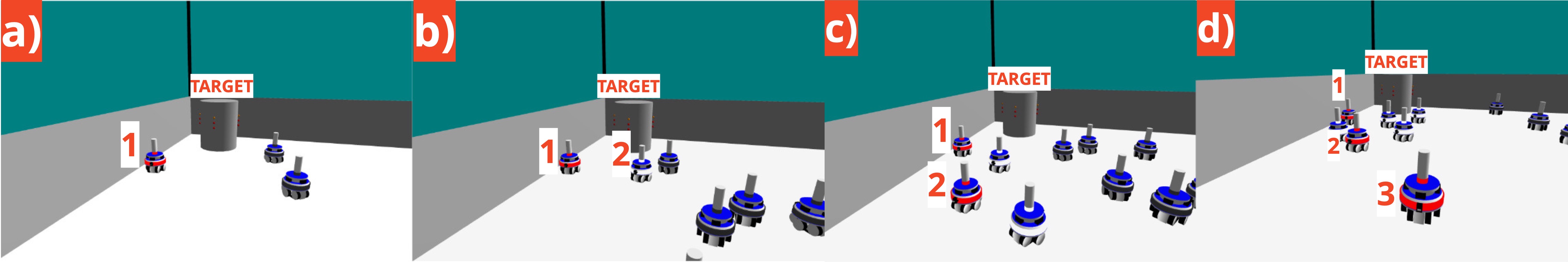}
    \caption{Sequential process of subgoal formation in multi-agent systems: (a) The first robot locates the goal and becomes a subgoal. (b) The second robot (white) locates the goal and seeks an appropriate position to become a subgoal. (c) The second white robot becomes a subgoal. (d) Three robots have successfully become subgoals.}
    \label{fig:subgoalprocess}
\end{figure}

\begin{figure}
    \centering
    \includegraphics[width=0.48\columnwidth]{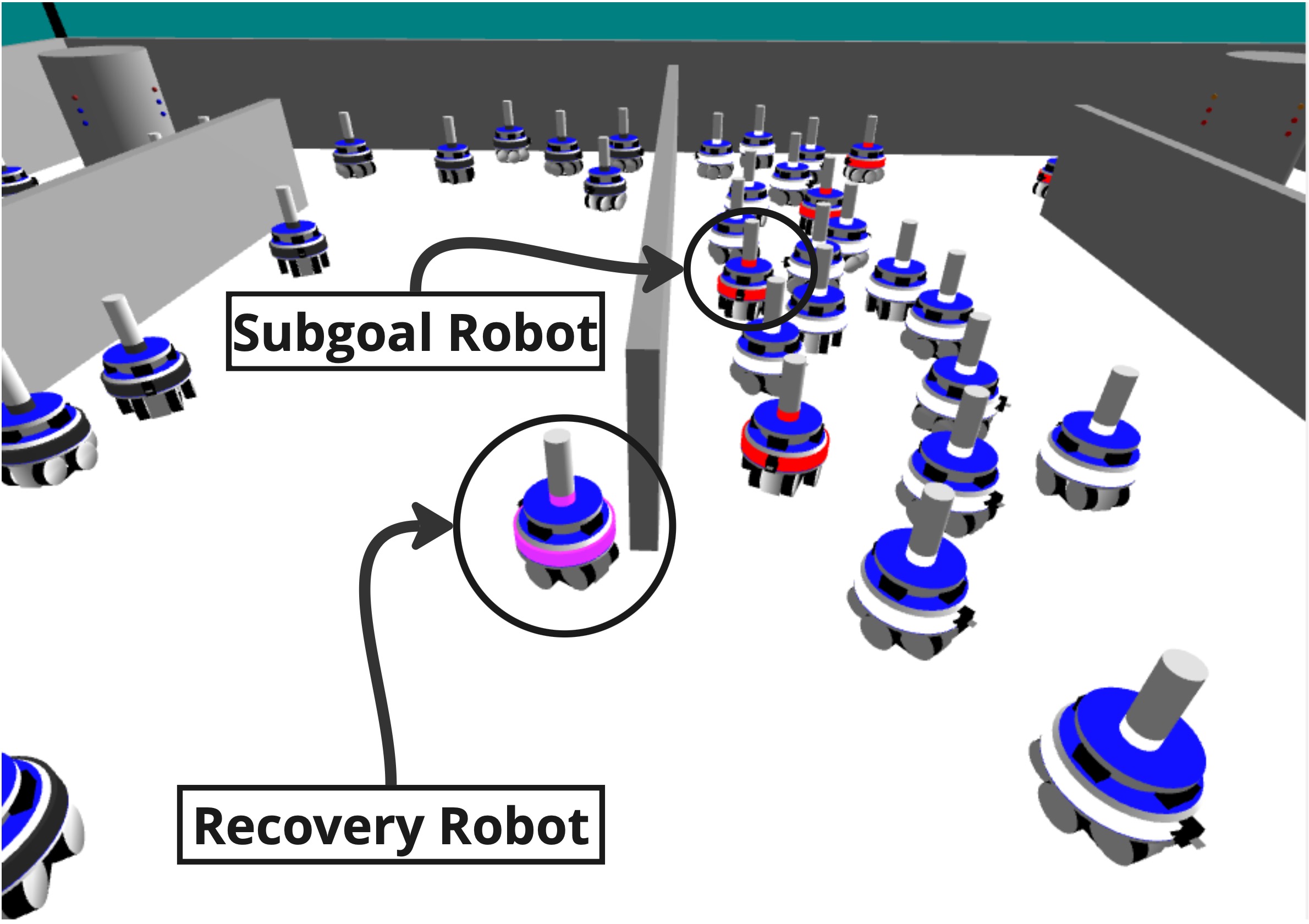}
    \caption{Recovery behavior and hidden location identification using purple illumination.}
    \label{fig:recoveryrobot}
\end{figure}

Additionally, a subgoal robot acts as a beacon, guiding other robots by emitting a color signal with its LED ring. Upon becoming a subgoal, it switches its LED to red. This triggers a distributed formation of one or more subgoal robots, continuing until a complete path is formed with intermediate subgoals from the goal back to the start. This subgoal formation process is systematically organized and documented in  Fig.~\ref{fig:subgoalprocess}.
One unique aspect of our research is the introduction of recovery robot behavior. Although robots can become subgoals within a range of 70 cm, there may be the cases where the goal/subgoal is not visible within that range. In such situations, the robot continues moving until it reaches the maximum visible range of 100 cm. If the robot loses visibility of the goal/subgoal within its visibility range, it assumes that there is an obstacle between the robot and the goal/subgoal. Consequently, the robot switches to the recovery robot state, as shown in Fig.~\ref{fig:recoveryrobot}. Recovery robots inform other robots to avoid entering the invisibility area. If a robot detects a recovery robot within a range of 20 cm, the recovery robot repulses it, ensuring that the robot avoids entering the blind spot. In certain cases, the robot may enter the repulsion range, triggering a state change and eventually becoming a subgoal.
\subsection{Path Formation Strategies}
\begin{figure}
    \centering
    \includegraphics[width=0.6\columnwidth]{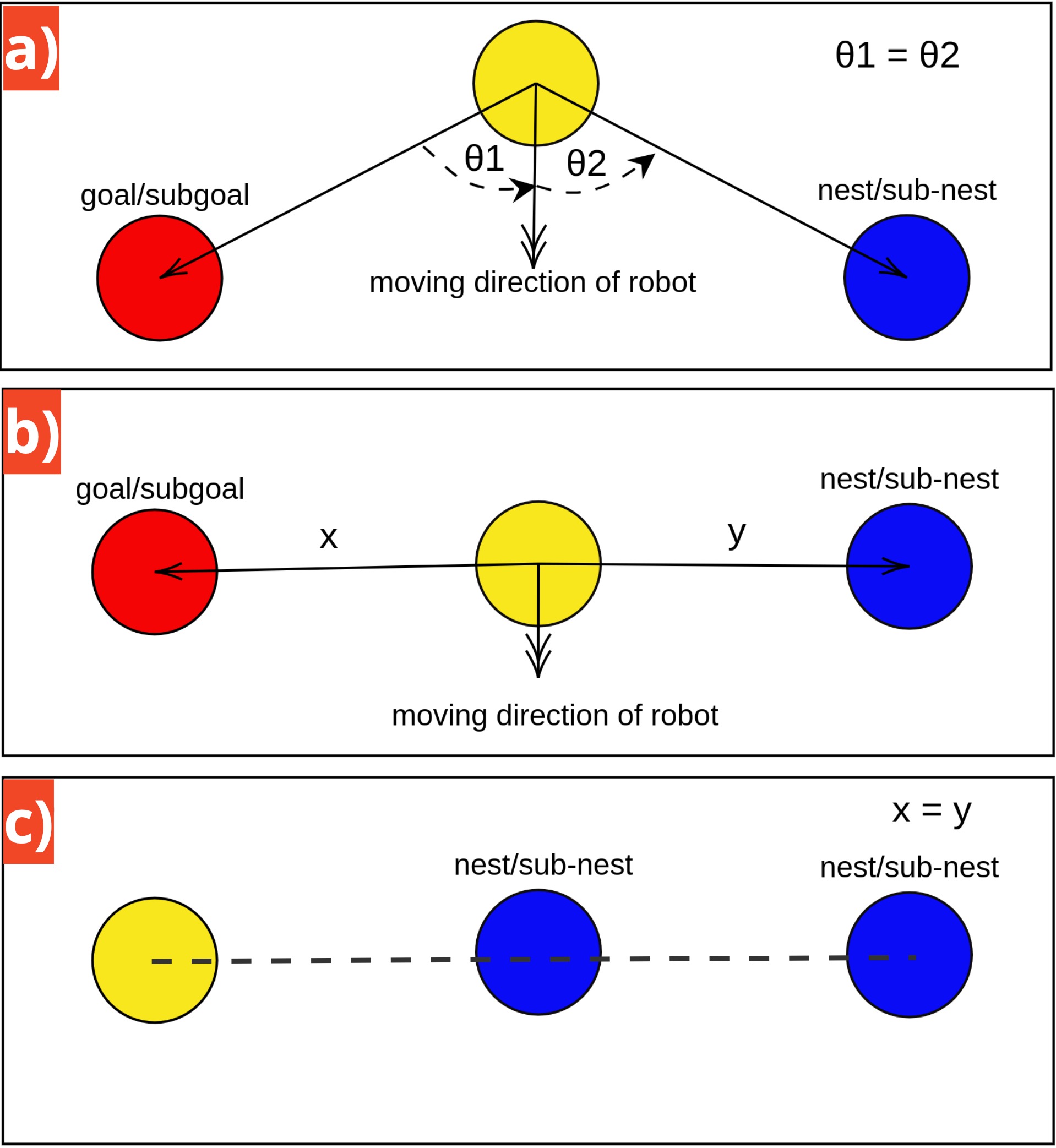}
    \caption{Heuristic optimization process.}
    \label{fig:optimization1}
\end{figure}

The path formation strategies in our approach involve two heuristic optimization processes: one from the starting point to the goal, and another from the goal to the starting point. The optimization process begins when a subgoal robot detects the nest (represented by the blue color). The first subgoal robot from the nest initiates the first alignment process with the second subgoal robot. Once this process is successfully completed, the first subgoal robot starts emitting the blue color, indicating that it is acting as a sub-nest. This process continues with subsequent subgoal robots until the last subgoal robot is reached. Four parameters are utilized in the first optimization strategy, as depicted in Fig.~\ref{fig:optimization1}: $\theta_1$ represents the goal/subgoal angle, $\theta_2$ represents the nest/sub-nest angle, while $x$ and $y$ denote the distances between the goal/subgoal and the processing robot, and between the nest/sub-nest and the processing robot, respectively. The first optimization process involves adjusting the robot's position to minimize the error angle.

\begin{figure*}[h!]
    \centering
    \begin{subfigure}[b]{0.22\textwidth}
        \includegraphics[width=\textwidth]{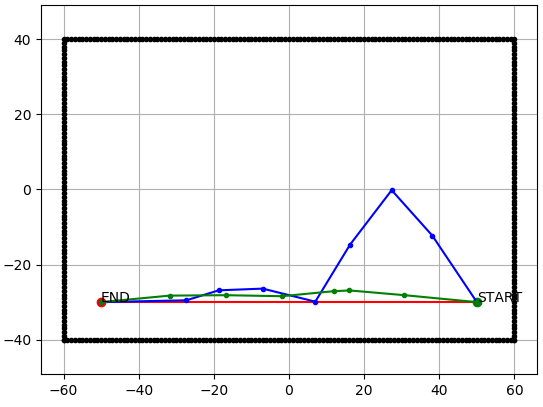}
        \caption{}
    \end{subfigure}
    \begin{subfigure}[b]{0.22\textwidth}
        \includegraphics[width=\textwidth]{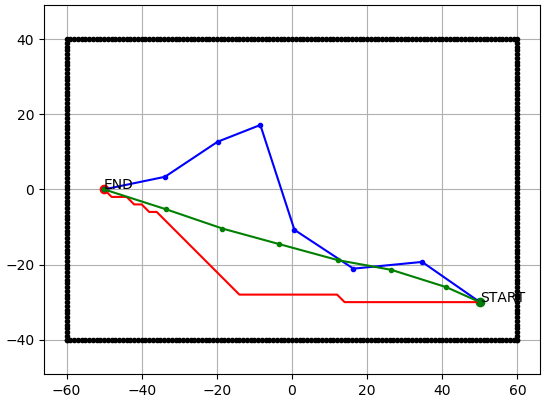}
        \caption{}
    \end{subfigure}
    \begin{subfigure}[b]{0.22\textwidth}
        \includegraphics[width=\textwidth]{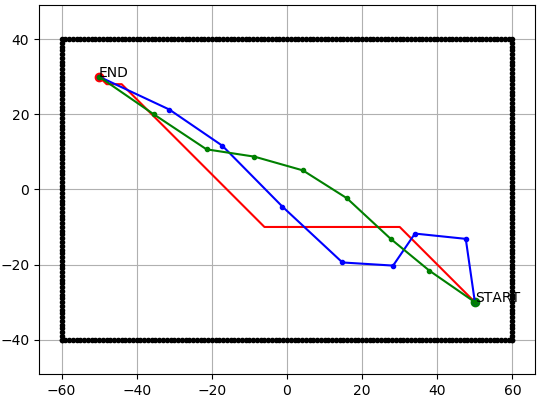}
        \caption{}
    \end{subfigure}
    \begin{subfigure}[b]{0.22\textwidth}
        \includegraphics[width=\textwidth]{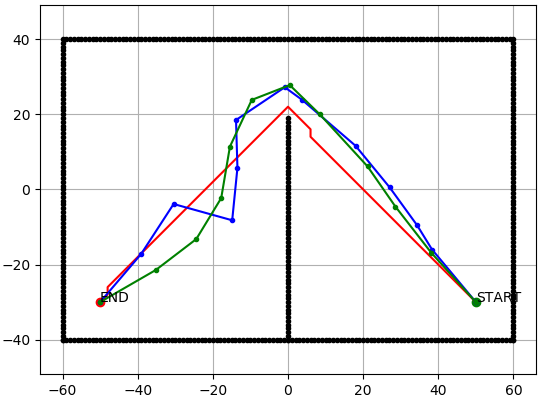}
        \caption{}
    \end{subfigure}
    \begin{subfigure}[b]{0.22\textwidth}
        \includegraphics[width=\textwidth]{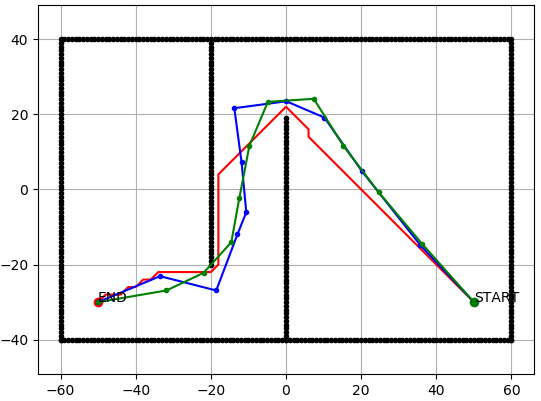}
        \caption{}
    \end{subfigure}
    \begin{subfigure}[b]{0.22\textwidth}
        \includegraphics[width=\textwidth]{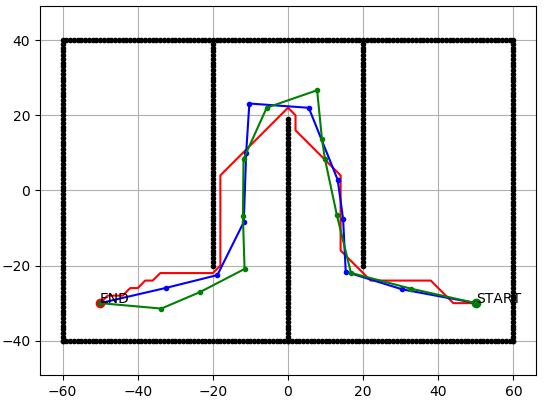}
        \caption{}
    \end{subfigure}
    \begin{subfigure}[b]{0.22\textwidth}
        \includegraphics[width=\textwidth]{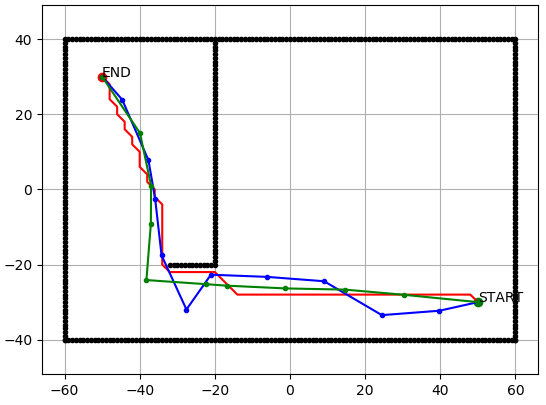}
        \caption{}
    \end{subfigure}
    \begin{subfigure}[b]{0.22\textwidth}
        \includegraphics[width=\textwidth]{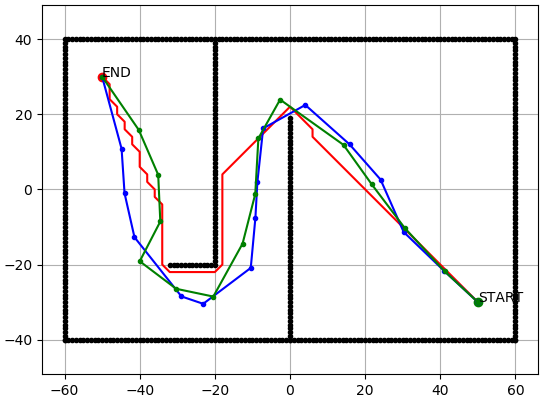}
        \caption{}
    \end{subfigure}
    \caption{Path planning for formation of 100 robots across various scenarios: (a-c) Robot trajectories in open environments. (d-f) Robot navigation in environments with obstacles. (g-h) Navigation strategies in complex environments with obstacles. Trajectories are color-coded: A* algorithm trajectories, subgoal formation trajectories, and optimized trajectories through task allocation strategies  are denoted by red, blue, and green colors, respectively.}
    \label{fig:all_robots}
\end{figure*}
Once the last subgoal robot completes the first alignment process, it proceeds to perform the second alignment process from the goal to the nest. It continues until it reaches the first subgoal robot from the starting position, similar to Fig.~\ref{fig:optimization1}. In cases where the alignment robot loses visibility of the subgoal or sub-nest within a certain visibility range during the first or second alignment process, it transitions to the recovery robot state. The role of the recovery robot is to inform other robots to avoid entering the invisibility area while they are in the subgoal formation process. Our control system has been designed to ensure the desired behavior described above to be achieved.
\subsection{Sensors and Actuators of Robots}
In the simulation environment, we utilize various actuators and sensors. Directional LEDs are controlled by the LED drivers, while the differential steering actuator manages robot wheel movements. The range-and-bearing actuator enables communication and location identification of message senders relative to receivers. This is complemented by a sensor that receives these messages. The foot bot proximity sensor generates a diffusion vector, and the light sensor measures light intensity $R$ as follows:
\begin{equation}
R=(I/x)^2 ,
\end{equation} 
 where $x$ is the distance to the light source and $I$ is the reference light intensity. This information helps create a potential field directed towards the nest. The positioning sensor tracks a robot location and orientation, aiding in guiding the resting robots back to their start point. An omnidirectional camera identifies colored blobs to form a color-based attraction vector.
\section{Evaluation and Comparisons}
\begin{figure}
    \centering
    \includegraphics[width=0.5\columnwidth]{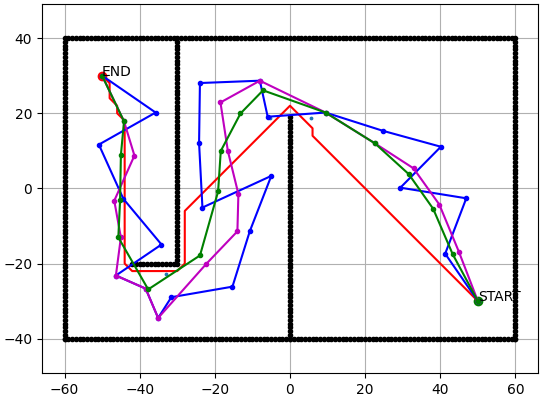}
    \caption{Path comparison: subgoal (blue), heuristic 1 (purple), heuristic 2 (green), A* algorithm (red).}
    \label{fig:comparison}
\end{figure}

The evaluation and comparison of the test model were performed in eight different environments, including three open environments, three obstacle environments, and two complex environments (see Fig.~\ref{fig:env_all}). Each environment was tested with varying robot number, ranging from 60 to 100 robots. Fig.~\ref{fig:all_robots} shows the comparison of formed paths. Fig.~\ref{fig:comparison} provides a visual comparison of heuristic optimization process.
\begin{table*}[t!]
\caption{Summary of time efficiency, path optimization, and resource utilization in multi-environment tests. }
\centering
\begin{tabular}{|c|c|c|c|c|c|c|} 
    \hline
    \multirow{2}{*}{Env.}  & \multicolumn{2}{|c|}{Time Taken (s)} & \multicolumn{3}{|c|}{Path length (m)} & \multirow{2}{*}{Resource Reduction (\%)} \\ 
    \cline{2-6}
    {} & Subgoal & Subgoal + Task allocation & A* & Subgoal & Subgoal + Task allocation & {} \\ 
    \hline
    1 & 4677 & 5047 & 5.00 & 6.57 & 5.015 & 77.1 \\ \hline
2 & 7348 & 2389 & 5.64 & 7.375 & 5.22 & 75.4 \\ \hline
3 & 5977 & 5026 & 6.645 & 7.475 & 5.93 & 75.4 \\ \hline
4 & 9754 & 8383 & 7.215 & 9.88 & 7.91 & 52.5 \\ \hline
5 & 10124 & 8164 & 7.945 & 7.515 & 7.31 & 49.2 \\ \hline
6 & 14471 & 15280 & 8.625 & 10.05 & 9.155 & 47.1 \\ \hline
7 & 10464 & 7679 & 7.155 & 7.115 & 7.295 & 76.2 \\ \hline
8 & 20123 & 19770 & 9.56 & 10.57 & 10.46 & 42.9 \\ \hline
\end{tabular}
\label{table:summary}
\end{table*}
The success rate of subgoal formation path formation was evaluated based on different environment types and robot sizes. The final path was compared with the subgoal path, optimization 1 path, optimization 2 path, and the A* algorithm path.
The performance of the task allocation model was also tested against the A* algorithm and the path before the implementation of the task allocation strategies. Table.~\ref{table:summary} presents the time taken to form the path in the default Argos step size, the path length in Argos default unit, and the percentage of resource reduction achieved in the eight different types of environments. The `Subgoal' column refers to the path formation model without improvement through task allocation, while the `Subgoal + Task allocation' column represents the path after incorporating the task allocation model. The evaluation revealed that without task allocation, 25\% of paths were shorter than those by the A* algorithm, with a success rate of 80\% over 40 test cases. The task allocation model demonstrated significant efficiency, reducing resource use by 61.93\% on average across tests. About 40\% of paths were shorter than those by A*, and all paths with task allocation were shorter than those without. Moreover, 87.5\% of paths with task allocation were formed faster compared to those without.

\section{Conclusion and Future Work}
This study developed a novel approach for swarm-based exploration and navigation, using behavior-based control strategies inspired by natural foraging. We implemented three main strategies: subgoal formation, heuristic alignment optimizations, and recovery behaviors, leading to a robust, adaptable path formation system. Our task allocation mechanism, which uses light signal interactions and structured communication protocols, made robot deployment costs go down by an average of 61.93\%. Comparative analysis with the A* algorithm revealed that our approach consistently achieved shorter paths in 40\% of test cases and faster formation in 87.5\% of test cases. Future work will focus on experiments with real robots in dynamic environments and integration of advanced communication protocols and Machine Learning techniques such as Deep Reinforcement Learning to optimize task allocation and achieve real-time response to environmental changes.

\section{Acknowledgements}
Research reported in this publication was financially supported by NeuroFleets (PVT) LTD.

\vspace{20pt}

\end{document}